\documentclass[11pt,a4paper]{article}
\usepackage[hyperref]{emnlp-ijcnlp-2019}
\usepackage{times}
\usepackage{latexsym}
\usepackage{graphicx}
\usepackage{amsfonts}
\usepackage{amsmath}
\usepackage{algorithm}
\usepackage[noend]{algorithmic}
\usepackage{natbib,subcaption}
\usepackage{multirow}
\usepackage{paralist}

\aclfinalcopy
\usepackage{url}
\usepackage{comment}
\usepackage{todonotes}

\newcommand{\SideNote}[2]{} 
\renewcommand{\SideNote}[2]{\todo[color=#1,size=\footnotesize]{#2}}

\title{Target Language-Aware Constrained Inference for Cross-lingual Dependency Parsing}  

\author{Tao Meng \\
  University of California, Los Angeles \\
  \texttt{tmeng@cs.ucla.edu} \\\And
  Nanyun Peng \\
  University of Southern California \\
  \texttt{npeng@isi.edu} \\ \AND
  Kai-Wei Chang \\
  University of California, Los Angeles \\
  \texttt{kw@kwchang.net} \\}

\date{}

\begin{document}
\maketitle
\begin{abstract}
    Prior work on cross-lingual dependency parsing often focuses on capturing the commonalities between source and target languages and overlooks the potential of leveraging linguistic properties of the languages to facilitate the transfer. 
    In this paper, we show that weak supervisions of linguistic knowledge for the target languages can improve a cross-lingual graph-based dependency parser substantially. 
    Specifically, we explore several types of \emph{corpus linguistic statistics} and compile them into \emph{corpus-wise constraints} to guide the inference process during the test time.
    We adapt two techniques, Lagrangian relaxation and posterior regularization, to conduct inference with corpus-statistics constraints. 
    Experiments show that the Lagrangian relaxation and posterior regularization inference improve the performances on 15 and 17 out of 19 target languages, respectively. 
    The improvements are especially significant for target languages that have different word order features from the source language. 
    
\end{abstract}

\section{Introduction}
    
    Natural language processing (NLP) techniques have achieved remarkable performance in a variety of tasks when sufficient training data is available. However, obtaining high-quality annotations for low-resource language tasks is challenging, and this poses great challenges to process low-resource languages. 
    To bridge the gap, cross-lingual transfer has been proposed to transfer models trained on high-resource languages (e.g., English) to low-resource languages (e.g., Tamil) to combat the resource scarcity problem. 
    Recent studies have demonstrated successes of transferring models across languages without retraining for NLP tasks, such as named entity recognition \citep{xie2018neural}, dependency parsing \citep{tiedemann2015cross,agic2014cross}, 
    and question answering \citep{joty2017cross}, using a shared multi-lingual word embedding space \citep{smith2017offline} or delexicalization approaches \citep{zeman2008cross,mcdonald2013universal}. 
    


    One key challenge for cross-lingual transfer is the \emph{differences} among languages; for example, languages may have different word orders. When transferring a model learned from a source language to target languages, the performance may drop significantly due to the differences. 
    To tackle this problem, various approaches have been proposed to better capture the \emph{commonalities} between the source and the target languages~\cite{mcdonald2011multi,guo2016representation,tackstrom2013target,agic2017cross,ahmad2018near}; however, they overlook the potential to leverage linguistic knowledge about the target language to account for the \emph{differences} between the source and the target languages
    to facilitate the transfer. 

    In this paper, we propose a complementary approach that studies how to leverage the linguistic knowledge about the target languages to help the transfer. Specifically, we use corpus linguistic statistics of the target languages as weak supervision signals to guide the test-time \emph{inference} process when parsing with a graph-based parser. 
    This approach is effective as the model only need to be trained \emph{once} on the source language and applied to many target languages using different constraints \emph{without retraining} the model.
    We argue that certain corpus linguistic statistics such as the word order (e.g., how often an adjective appears before or after a noun) can be easily obtained from available resources such as World Atlas of Language Structures (WALS)~\cite{wals}. 
    To incorporate the corpus linguistic statistics to a cross-lingual parser, we compile them into corpus-wise constraints and adopt two families of methods: 1) Lagrangian relaxation (LR) and 2) posterior regularization (PR) to solve the constrained inference problem. 
    The algorithms take the original graph-based parsing inference as a sub-routine, and LR iteratively adjusts the pair-wise potentials until the constraints are (loosely) satisfied, while PR finds a feasible distribution and do inference based on that. 
    The constrained inference framework is general and supports any 
    knowledge that can be formulated as a first-order logic \citep{roth2004linear}. 
    
    We evaluate the proposed approach under the single-source transfer setting using English as the source language and test on 19 target languages covering a broad range of language families with low-resource languages such as Tamil and Welsh. We demonstrate that by adding three simple corpus-wise constraints derived from WALS features, the performances improve in 15 and 17 out of 19 languages when using Lagrangian relaxation and posterior regularization techniques, respectively. 
    The improvements are especially substantial when the target language features are distant from the source language. For example, our framework improves the UAS score of Urdu by 15.7\%, and Tamil by 7.3\%.\footnote{The code and data are available at \url{https://github.com/MtSomeThree/CrossLingualDependencyParsing.}}  
\section{Constrained Cross-Lingual Parsing}      
\begin{figure}[t]
            \centering
            \includegraphics[width=\linewidth]{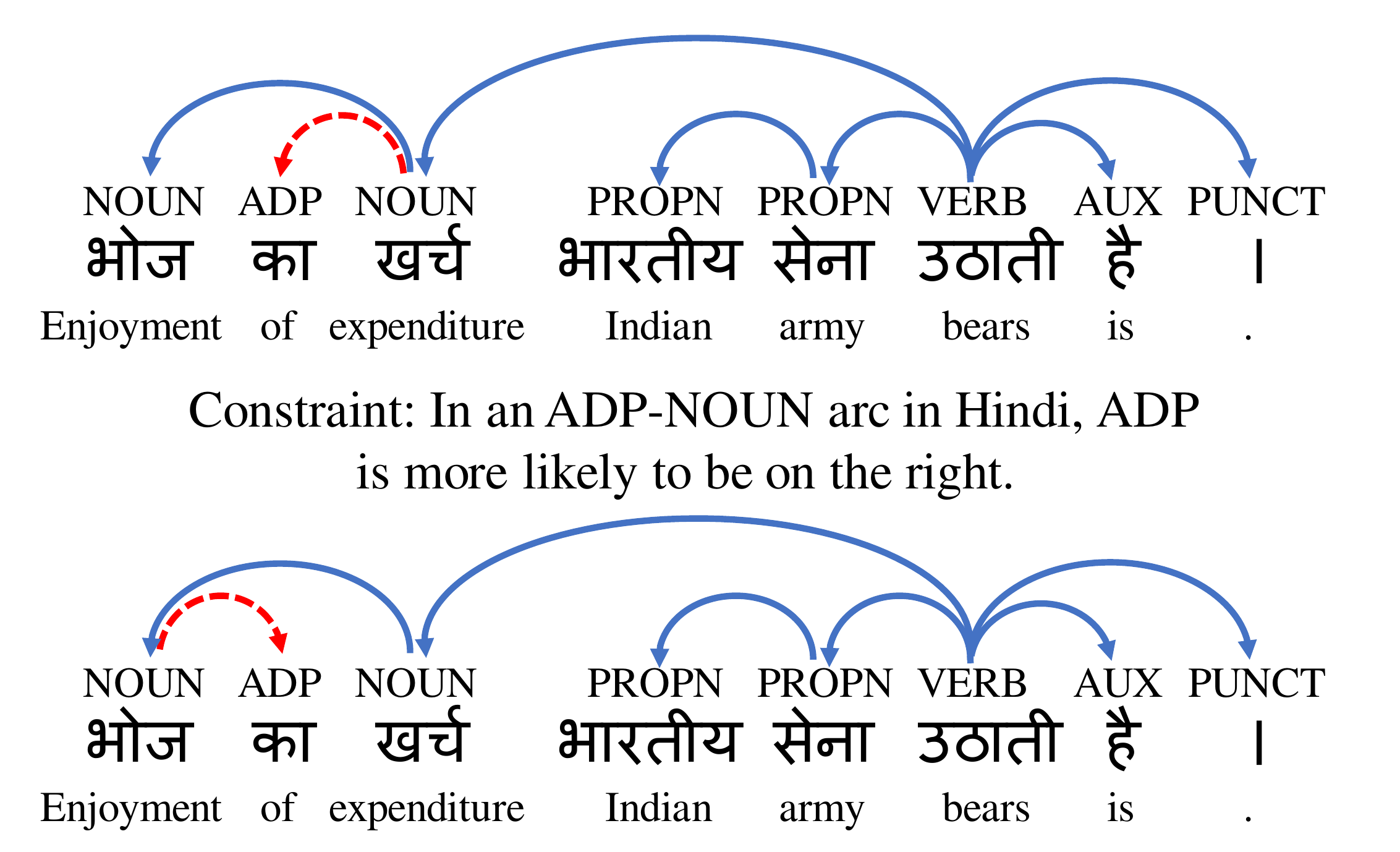}  %
            \caption{An running example of Hindi. On the top there is the inference result of a baseline model trained on English. In English, ADP is mostly on the left of the NOUN, so the potential of the correct ADP-NOUN arc is lower. With the help of corpus-statistics constraints, the potential is adjusted and the model gets correct inference result as shown in the bottom. The dashed lines highlight the difference.}
            \label{fig:constraint}
        \end{figure}

    Our work focuses on the graph-based dependency parser~\citep{McDonald:2005} in the zero-shot single-source transfer setting as in \newcite{ahmad2018near}. However, the proposed algorithms can be extended other transfer settings. 
    Given a trained model, we derive corpus-statistics constraints and apply them to correct errors caused by word order differences between the source and the target language \emph{during the inference time}. Figure \ref{fig:constraint} 
    shows an example of how constraints can influence the inference results. 
    
    In this section, we first give a quick review of the graph-based parser and introduce the notations. We then discuss how to formulate corpus-wise constraints based on corpus linguistic statistics for guiding the graph-based parser. 
    
\subsection{Background: Graph-Based Parser}
\label{sec:background}
    A graph-based parser learns a scoring function for every pair of words in a sentence and conducts inference to derive a directed spanning tree with the highest accumulated score. Formally, given the $k$-th sentence $\mathbf{w}_k=(w_{k1},\ldots,w_{kL(k)})$ where $L(k)$ denotes the length of the $k$-th sentence, a graph-based parser learns a score matrix $S^{(k)}$, where $S^{(k)}_{ij}$ denotes the score to form an arc from word $w_{ki}$ to word $w_{kj}$. Let $y_k$ be an indicator function that $y_k(i,j)\in\{0,1\}$ denotes the arc from $w_{ki}$ to $w_{kj}$. The maximum directed spanning tree inference can be formulated as an integer linear programming (ILP) problem:
    \begin{equation}
    \label{eq:ilp}
    y^*_k=\arg\max_{y_k\in\mathcal{Y}_k} \sum_{i,j} S^{(k)}_{ij}y_k(i,j),
    \end{equation}
    where 
    $\mathcal{Y}_k$ is the set of legal dependency trees of sentence $k$. In recent years, neural network approaches~\citep{kiperwasser2016simple,wang2016graph, kuncoro2016distilling,dozat2017biaffiine} have been applied to modeling the scoring matrix $S^{(k)}$ and have achieved great performance in dependency parsing.
    
    From the probabilistic point of view, if we assume for different $i, j$, the edge probabilities $P(y_k(i,j)=1|\mathbf{w}_k)$ are mutually conditional independent, the probability of a whole parse tree can be written as
    \begin{equation}
    \label{eq:prob}
    P(y_k|\mathbf{w}_k)=\prod_{i,j}P(y_k(i,j)=1|\mathbf{w}_k)^{y_k(i,j)}.
    \end{equation}
    If we set 
    $S^{(k)}_{ij}=\log P(y_k(i,j)=1|\mathbf{w}_k)+Z'_j,$
    where $Z'_j$ is a constant term, then Eq. \eqref{eq:ilp} can be regarded as the following maximum a posteriori (MAP) inference problem:
    \begin{equation} \label{eq:map}
    \begin{split}
      y^*_k&=\arg\max_{y_k\in\mathcal{Y}_k} P(y_k|\mathbf{w}_k) \\
      &= \arg\max_{y_k\in\mathcal{Y}_k} \sum_{i,j} \log P(y_k(i,j)=1|\mathbf{w}_k)  y_k(i,j).
    \end{split}
    \end{equation}
    

    \subsection{Corpus-Wise Constraints} \label{sec:constraintsF}
        Given the inference problems as in equations~\eqref{eq:ilp} and~\eqref{eq:map}, additional constraints can be imposed to incorporate expert knowledge about the languages to help yield a better parser. 
        Instance-level constraints have been explored in the literature of dependency parsing, both in the monolingual \citep{dryer2007word} and cross-lingual transfer \citep{tackstrom2013target} settings.
        However, most word order features for a language are non-deterministic and cannot be compiled into instance-wise constraints. 
        
        In this work, we introduce corpus-wise constraints to leverage the non-deterministic features for cross-lingual parser. We compile the following two types of corpus-wise constraints based on corpus linguistics statistics:
        \begin{compactitem}
        \item \emph{Unary} constraints consider statistics regarding a particular POS tag $(POS)$. 
        \item \emph{Binary} constraints consider statistics regarding a pair of POS tags $(POS_1,POS_2)$.
        \end{compactitem} 
        Specifically, a unary constraint specifies the ratio $r$ of the heads of a particular $POS$ appears on the left of that $POS$.\footnote{The ratio for the head being on the right of that $POS$ is thereby $1-r$.}
        Similarly, a binary constraint specifies the ratio $r$ of $POS_1$ being on the left of $POS_2$ when there is an arc between $POS_1$ and $POS_2$.  
        
        The ratios $r$ for the constraints are called corpus statistics, which can be estimated in one of the following ways: a) leveraging existing linguistics resources or consulting linguists; b) leveraging a higher-resource language that is similar to the target language (e.g., Finnish and Estonian) to collect the statistics. 
        In this paper, we explore the first option and leverage the WALS features, which provide a reference for word order typology, to estimate the ratios.
        
        \paragraph{Compile Constraints From WALS Features.}
        For a particular language, once we collect the corpus-statistics of a pair of POS tags, we can formulate a binary constraint. There are different ways to estimate the corpus-statistics. For example, \citet{ostling2015word} utilizes a small amount of parallel data to estimate the dominant word orders. In this paper, we simply utilize a small subset of WALS features 
        that show the dominant order of some POS pairs (e.g. adjective and noun) in a language. They can be directly compiled into binary constraints.
        
        Similarly, we can estimate the ratio for unary constraints based on WALS features. For a particular POS tag, we choose all WALS features related to it to formulate a feature vector $f$. The mapping from the vector $f$ to the unary constraint ratio $r$ is learnable: for each language with annotated data, we can get a WALS feature vector $f_{lang}$ and a ratio $r_{lang}$ from the annotation. We only need a small amount of data to estimate $r_{lang}$ well. Given a set of languages with feature vectors and estimated ratios, we can learn the mapping by a simple linear regression, and apply it to estimate the ratio of any target language to compile a unary constraint.  
        
        \subsection{Formulate Constraints}
        In the following, we mathematically formulate the corpus-wise constraints. Note that these constraints are based on the statistics over the entire corpus. 
        For a unary constraint $C_u^{(POS)}$, let $P$ denotes a set of word with part-of-speech tag $POS$. We define $C_u^+:=\{(k,i,j)|w_{kj}\in P\wedge i<j\}$ as the set of arcs where the head of word in $P$ is on its left and $C_u^-:=\{(k,i,j)|w_{kj}\in P\wedge i>j\}$, conversely. 
    
        For a binary constraint $C_b^{(POS_1, POS_2)}$, we denotes $P_1$, $P_2$ as a set of word with part-of-speech tag $POS_1$, $POS_2$, respectively. We then define
        $C_b^+$ as the set of arcs with two ends $w_{ki}\in P_1$ and $w_{kj} \in P_2$, and $w_{ki}$ is on the left of $w_{kj}$. We define $C_b^-$ Similarly.
        Formally, 
        \begin{equation*}
            \begin{split}
            C_b^+:=&\{(k,i,j)|w_{ki}\in P_1\wedge w_{kj}\in P_2\wedge i<j\}\\
                &\cup\{(k,j,i)|w_{ki}\in P_1\wedge w_{kj}\in P_2\wedge i<j\},\\
            C_b^-:=&\{(k,i,j)|w_{ki}\in P_1\wedge w_{kj}\in P_2\wedge i>j\}\\
                &\cup\{(k,j,i)|w_{ki}\in P_1\wedge w_{kj}\in P_2\wedge i>j\}.
            \end{split}
        \end{equation*}

        For notational simplicity, we use $C$ to denote all constraints including both the unary and binary ones. The ratio function $R(C,Y)$ for a constraint $C$ given the parse trees $Y$ can be defined as:
        \begin{equation*}
            \label{eq:const}
             R(C,Y)\!\!=\!\!\frac{\sum\nolimits_k\sum\nolimits_{(i,j):(k,i,j)\in C^+}y_k(i,j)}{\sum\nolimits_k\sum\nolimits_{(i,j):(k,i,j)\in C^+ \cup C^-}y_k(i,j)}. 
       \end{equation*}
       We want to enforce the ratio $R(C,Y)$ estimated from $Y$ to be consistent with a value $r$ (see Sec. \ref{sec:constraintsF}), which formulates a constraint 
       \begin{equation*}
              r-\theta \leq R(C,Y) \leq r+\theta
       \end{equation*}
     where $\theta$ is a tolerance margin. Note that the instance-level hard constraint is a special case of the corpus-statistics constraint when $r=0$ or $r=1$. 
        
        Given a set of corpus-statistics constraints  $\mathcal{C}=\{C_1,C_2,\ldots,C_n\}$ with corresponding corpus statistics $\mathbf{r}=\{r_1,r_2,\ldots,r_n\}$, the objective of the constrained inference is:
        \begin{equation}
        \label{eq:ilpc}
            \begin{split}
                \max_{Y\in\mathcal{Y}} \quad  & \sum\nolimits_k \sum\nolimits_{i,j\leq L(k)} S^{(k)}_{ij}y_k(i,j), \\
            \mbox{s.t.}   \quad & r_i\!-\!\theta_{i}\leq R(C_i,Y)\!\leq\!r_i\!+\!\theta_{i}, i\in [N],
            \end{split}
        \end{equation}
        where $\mathcal{Y}$ denotes the set of all possible dependency trees. 
        As all the constraints can be written as a linear inequality with respect to $y_k(i,j)$. Eq. \eqref{eq:ilpc} is an ILP.

\section{Inference with Corpus-Statistics Constraints} 
\begin{figure}[t]
        \centering
        \includegraphics[width=.8\linewidth]{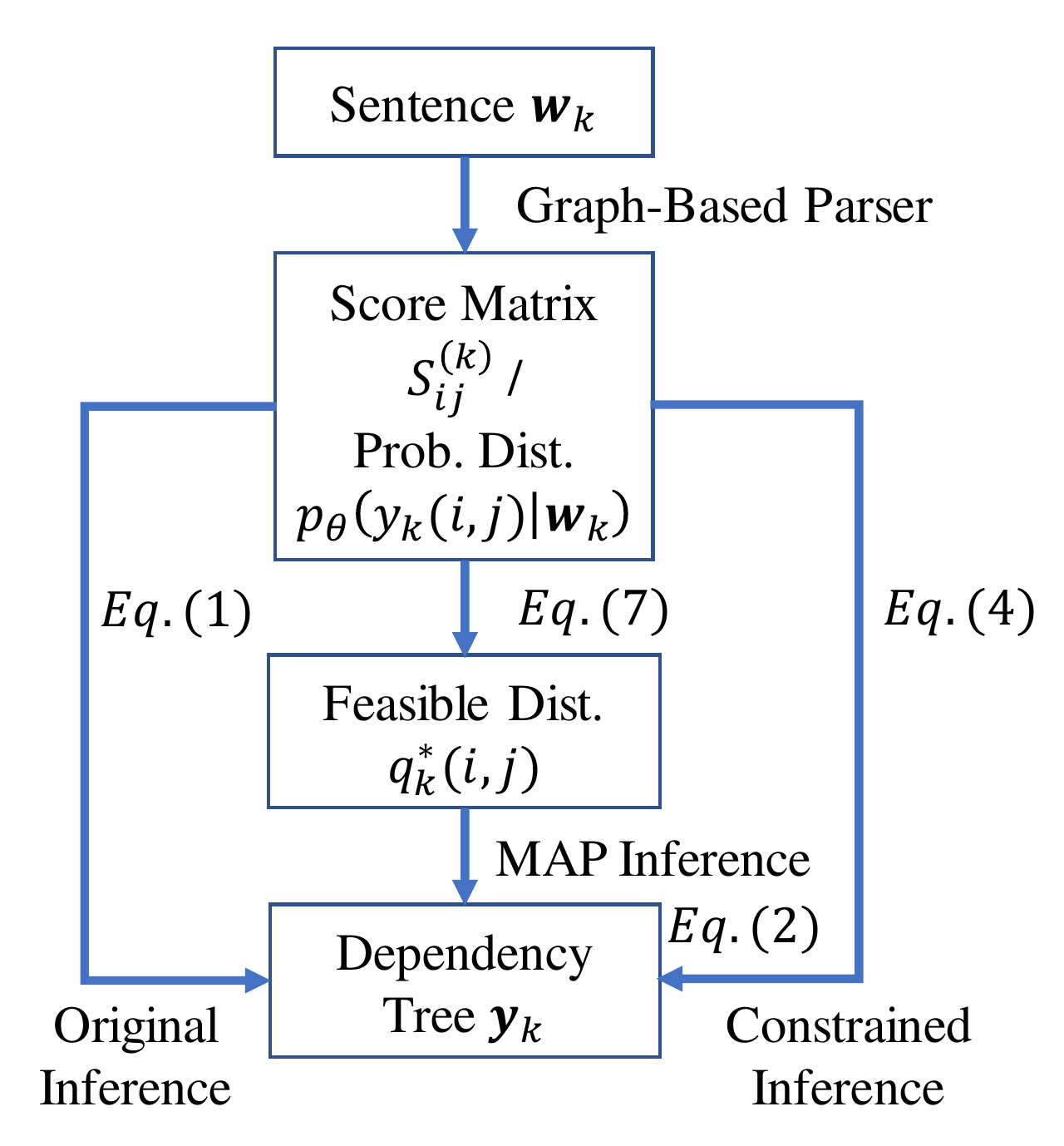}  %
        \caption{The pipelines of the baseline method (left), Lagrangian relaxation (right) and posterior regularization (middle). Lagrangian relaxation converts constrained inference to an unconstrained optimization problem using Lagrange's method. Posterior regularization method is working on the distribution space. For a given distribution, PR finds the closest feasible distribution and conduct MAP inference.}
        \label{fig:method}
        \vspace{-1.5em}
\end{figure}
   The ILP problem in Eq.~\eqref{eq:ilpc} is in general an NP-hard problem; especially, it involves variables associated with the entire corpus. Without constraints, Eq. \eqref{eq:ilpc} can be decoupled into $K$ sub-problems, and the inference with respect to each sentence can be solved independently as in Eq. \eqref{eq:ilp}. In this way, an efficient inference algorithm such as maximum directed spanning tree algorithm~\cite{chu-liu-1965} can be used.
        
    However, with the corpus-wise constraints, directly solving Eq. \eqref{eq:ilpc} is infeasible.
    Therefore, we explore two algorithms for inference with corpus-statistics constraints: Lagrangian relaxation and posterior regularization \citep{ganchev2010posterior}. 
    The Lagrangian relaxation algorithm introduces Lagrangian multipliers to relax the constraint optimization problem to an unconstrained optimization problem, and estimates the  Lagrangian multipliers with gradient-based methods.
    The posterior regularization algorithm uses the constraints of the target language to define a feasible set of parse tree distributions, and find a feasible distribution that is closest to the parse tree distribution trained on the source language by minimizing the KL-divergence. The constrained inference problem can then be converted into an MAP inference problem on the best feasible distribution. 
    Figure \ref{fig:method} illustrates the procedure of the original inference, Lagrangian relaxation, and posterior regularization. 

    \subsection{Lagrangian Relaxation}

        Lagrangian relaxation has been applied in various NLP applications~\citep{rush2012tutorial,rush2011exact}. 
        In Eq. \eqref{eq:ilpc}, each constraint $C_i$ involves two inequality constraints: $ R(C_i,Y)-r_i+\theta\geq 0,$ and $ r_i+\theta-R(C_i,Y)\geq 0.$
    Instead of treating these two constraints separately, we consider a heuristic to optimize with equality constraints $R(C_i,Y) = r_i, i\in [N]$
            and terminate earlier when constraints in Eq. \eqref{eq:ilpc} are satisfied. Despite this approach does not guarantee the solution is optimal if all the constraints are satisfied as the original Lagrangian relaxation algorithm does, in practice, the inference converges faster (as the number of Lagrangian multipliers is half) and the parsing performance maintains.

        In the following, we derive the constrained inference algorithm for corpus-statistics constraints. 
        First, we rewrite the equality constraint $R(C,Y)=r$ by substituting $R(C, Y)$ with Eq.~\eqref{eq:const}:
        \begin{eqnarray} \label{eq:lr_const}
            & (1-r)\sum\limits_k\sum\limits_{(i,j):(k,i,j)\in C^+_i}y_k(i,j) \nonumber\\
            & -r\sum\limits_k\sum\limits_{(i,j):(k,i,j)\in C^-_i}y_k(i,j)=0, 
        \end{eqnarray}
        We use F(C) to denote the left-hand-side of Eq.~\eqref{eq:lr_const}, which is linear w.r.t. $y_k$. 
        Then, the Lagrangian relaxation of the constrained inference problem can be written as:
        \begin{equation*}
        \begin{split}
         L(Y,\lambda;\mathcal{C}) 
         &= \sum_k \sum_{i,j} S^{(k)}_{ij}y_k(i,j) \!+\!  \sum\limits_{i\in [N]} \lambda_i F(C_i),
         \end{split}
       \end{equation*}
        where $\lambda_i$ is called Lagrangian multiplier. 
        It is well-known that we can solve the dual form of the constrained inference problem:  
        $$\max\limits_{Y\in\mathcal{Y}}\min\limits_{\lambda\geq 0}L(Y,\lambda;\mathcal{C}).$$
        To solve the dual form, we initialize  $\lambda_i$ to be 0. At iteration $t$, we firstly conduct an constraint-augmented inference with a fixed $\lambda^{(t)}$:
         \begin{equation}
         \label{eq:cinfer}
         \hat{Y}^{(t)}=\max\limits_{Y\in\mathcal{Y}}L(Y,\lambda^{(t)};\mathcal{C}).
         \end{equation}

               
            As $F(C)$ is a linear function w.r.t $y_k$, we combine it with $S^{(k)}_{ij}y_k(i,j)$. In this way, the inference problem Eq. \eqref{eq:cinfer} can be treated as a special case of Eq. \eqref{eq:ilp} with a different scoring matrix $S^{(k)}$.   
            In this way, we can treat the inference on every sentence independently and leverage existing inference techniques. 
            
            
            \fi    
            
        After solving the constraint-augmented inference, we compute the ratio of every constraint $\hat{r_i}^{(t)}=R(C_i,\hat{Y}^{(t)}),$
        and use gradient ascent algorithm to update the Lagrangian multipliers
        $$\lambda_i^{(t+1)}=\lambda_i^{(t)}+\alpha^{(t)}(r_i-\hat{r_i}).$$
        Here $\alpha^{(t)}$ denote the step size at iteration $t$. The algorithm is shown in Algorithm \ref{alg:A}.
        
        \begin{algorithm}[t]
            \caption{\label{algorithm1} Lagrangian Relaxation for Constraint Inference}
            \label{alg:A}
            \hspace*{0.02in} 
            {\bf Input:} Constraints $\mathcal{C}=\{C_i\}_{i=1}^N$, corresponding ratio $\mathbf{r}=\{r_i\}_{i=1}^N$, tolerance margin $\mathbf{\theta}=\{\theta_i\}_{i=1}^N$,  learning rate decay $\eta$,  initial learning rate $\alpha_0$\\
            \hspace*{0.02in} {\bf Output:} parse trees $\hat{Y}$
            \begin{algorithmic}[1]
                \STATE $\alpha \leftarrow \alpha_0$
                \STATE $\lambda_i \leftarrow 0,\ i\in [N]$
                \REPEAT
                    \STATE $\hat{Y}\leftarrow$  $\arg\max_{Y\in\mathcal{Y}}L(Y,\lambda;\mathcal{C})$
                    \STATE $\hat{r_i}\leftarrow R(C_i,\hat{Y}),\ i\in[N]$
                    \IF {$\forall i, \mathop{abs}(r_i-\hat{r_i}) \leq \theta$}
                    \RETURN $\hat{Y}$ 
                    \ENDIF
                    \STATE $\lambda_i\leftarrow \alpha (\hat{r_i}-r_i),\ i\in[N]$
                    \STATE $\alpha\leftarrow \eta\alpha$
                \UNTIL{MAX\_ITER times}
                \RETURN $\hat{Y}$
            \end{algorithmic}
        \end{algorithm}
        
  
    \subsection{Posterior Regularization}
    From a probabilistic point of view, the parser model learns parameters $\theta$ to realize Eq. \eqref{eq:prob}. 
    During the inference, the model predict a probability distribution $p_\theta(\mathbf{Y|W})$ over possible parse trees given a sentence. 
    The posterior regularization algorithm first defines a feasible set of the probability distributions w.r.t the given constraints, and looks for the closest feasible distribution $q^*(\mathbf{Y})$ to the model distribution $p_\theta(\mathbf{Y}|\mathbf{W})$. 
    The best parse tree is given by $\arg\max_{\mathbf{Y}}q^*(\mathbf{Y})$. Specifically, we define the feasible set as: 
    \begin{equation*}
        \label{eq:Qset}
            Q\!=\!\{q(\mathbf{Y})|r_i-\theta_i\!\leq\!R(C_i,q(\mathbf{Y}))\!\leq\!r_i+\theta_i,i\!\in\![N]\},
    \end{equation*}
    where $R(C,q)\!=\!\frac{\sum_k\sum_{(i,j):(k,i,j)\in C^+}q_k(i,j)}{\sum_k\sum_{(i,j):(k,i,j)\in C^+ \cup C^-}q_k(i,j)},$
    with
    $q_k(i,j)=E_{Y\sim q(Y)}[y_k(i,j)]$.
    
    To measure the distance between two distribution, we use the KL-divergence, and find the best feasible distribution $q^*(\mathbf{Y})$:
    \begin{equation}
        \label{eq:qqq}
        q^*(\mathbf{Y})=\arg\min_{q\in Q} KL(q(\mathbf{Y})\|p_\theta(\mathbf{Y|W})). 
    \end{equation}
    
    If the feasible set has the expectation form:
    \begin{equation}
    \label{eq:Q}
    \{q|\mathbb{E}_{\mathbf{Y}\sim q} [\phi(\mathbf{Y})]\leq \mathbf{b}\},
    \end{equation}
    Eq. \eqref{eq:qqq} has a simple close form solution~\cite{ganchev2010posterior}:
    \begin{equation}
        \label{eq:solution}
        q^*(\mathbf{Y})=\frac{p_\theta(\mathbf{Y}|\mathbf{W})\exp(-\lambda^*\cdot\phi(\mathbf{Y}))}{Z(\lambda^*)},
    \end{equation}
    where $\lambda^*$ is the solution of
    \begin{equation}
        \label{eq:lambda}
        \begin{aligned}
            \lambda^*\!&=\!\arg\max\limits_{\lambda\geq 0}-\mathbf{b}\cdot\lambda-\log Z(\lambda),\\
            Z(\lambda)\!&=\!\sum\limits_{\mathbf{Y}'}p_\theta(\mathbf{Y}'|\mathbf{W})\exp(-\lambda^*\cdot\phi(\mathbf{Y}')).
        \end{aligned}
    \end{equation}
    
    In the rest of this section, we first show that the feasible set $Q$ we considered above can be reformulated in the form of Eq. \eqref{eq:Q}, and then we discuss how to solve Eq. \eqref{eq:lambda}. To show that the inequality $R(C,q(\mathbf{Y}))\leq r,$ in $Q$ 
    can be formulated in the form of Eq. \eqref{eq:Q}, we set
    $$\phi_C(\mathbf{Y})=\sum_{k,i,j}y_k(i,j)\phi_C(k,i,j),$$ and 
    \begin{equation}
        \label{eq:phi}
        \phi_C(k,i,j)\!=\!
        \begin{cases}
            1-r & (k,i,j) \!\in\! C^+\\
            -r & (k,i,j) \!\in\! C^-\\
            0 & (k,i,j) \!\notin\! C^+ \!\cup\! C^-.\\
        \end{cases}
    \end{equation}
    Similarly, we can derive $R(C,q(\mathbf{Y}))\geq r,$ into the same form and rewrite $Q$ as
    \begin{equation*}
        \label{eq:pr}
        Q\!=\!\{q(\mathbf{Y})\ |\ \mathbb{E}_{\mathbf{Y}\sim q}\phi(\mathbf{Y})\leq \mathbf{0}\},
    \end{equation*}
    where $\phi=(\phi_{C_1},\phi_{C_2},...,\phi_{C_N})$ is a collection of the constraints. The detailed derivations can be found in Appendix~\ref{app-sec:details}.

    We solve Eq. \eqref{eq:lambda} by sub-gradient descent\footnote{In implementation, we use stochastic gradient descent with Adam optimizer~\cite{kingma2014adam}}. Noting that there can be exponential number of terms in $Z(\lambda)$, we firstly need to factorize $Z(\lambda)$ from corpus level to instance level and arc level, and compute the gradient. The technical details are in  Appendix~\ref{app-sec:details}. With the optimal $\lambda^*$, we can compute the feasible distribution $q^*(\mathbf{Y})$ given $p_\theta$ and $\mathbf{W}$. Noting that the solution Eq. \eqref{eq:solution} can also be factorized to arc-level:
    \begin{equation}
        \label{eq:arc_solution}
        q^*_k(i,j)\propto p_\theta(y_k(i,j)|\mathbf{w_k})\exp(-\lambda^*\cdot\phi(k,i,j)),
    \end{equation}
    here $q^*_k(i,j)$ denote the arc-level distribution $q^*(y_k(i,j)=1)$ satisfying Eq. \eqref{eq:map}. We then do MAP inference based on $q$, which is actually a minimal spanning tree problem same as before. 
    Algorithm \ref{alg:B} summarizes the process. 
    
    \begin{algorithm}[t]
            \caption{\label{algorithm2} Posterior Regularization}
            \label{alg:B}
            \hspace*{0.02in} 
            {\bf Input:} Constraints $\mathcal{C}=\{C_i\}_{i=1}^N$, corresponding ratio $\mathbf{r}=\{r_i\}_{i=1}^N$, tolerance margin $\mathbf{\theta}=\{\theta_i\}_{i=1}^N$\\
            \hspace*{0.02in} {\bf Output:} parse trees $\hat{Y}$
            \begin{algorithmic}[1]
                \STATE $p_\theta(y_k(i,j)=1)$ $\leftarrow$ normalize$\left(\exp\left(S^{(k)}_{ij}\right)\right)$
                \STATE $\phi$ $\leftarrow$ defined by Eq. \eqref{eq:phi}
                \STATE $\lambda_i \leftarrow 0,\ i\in [N]$
                \REPEAT
                    \STATE estimate $\mathbf{g}\leftarrow \partial \log Z(\lambda)/\partial \lambda$ in a batch
                    \STATE update $\lambda$ based on $\mathbf{g}$
                \UNTIL{MAX\_ITER times}
                \FOR{each $(k,i,j)$}
                    \STATE $q_k(i,j) \leftarrow$ defined by Eq. \eqref{eq:arc_solution} 
                \ENDFOR
                \STATE $\hat{Y}\leftarrow$ MAP inference based on $q    $
                \RETURN $\hat{Y}$
            \end{algorithmic}
        \end{algorithm}

\section{Experiments}
In this section, we evaluate the proposed algorithms by transferring an English dependency parser to 19 target languages covering 13 language families of real low-resource languages. We first introduce the experimental setup including data selection and constraint details and then discuss the results as well as in-depth analysis.
         \begin{table*}[t!]
             \begin{tabular}{@{ }l@{\ \ } |@{\ \ }c@{ }| @{\ \ }c@{ } | @{\ \ }c@{ } | c@{\  \ }c@{\ \ }c | c@{\ \ }c@{\ \ }c@{}}
                \hline
                \multirow{2}{*}{\textbf{Family}} & \multirow{2}{*}{\textbf{Lang.}}  & \multirow{2}{*}{\textbf{Features}} & \multirow{2}{*}{\textbf{Baseline}}  & \multicolumn{3}{c|}{\textbf{Lagrangian Relaxation}} & \multicolumn{3}{c}{\textbf{Posterior Regularization}}\\
                \cline{5-10}
                & & & & \textbf{Oracle} & \textbf{WALS}  & $\Delta$\textbf{WALS} & \textbf{Oracle} & \textbf{WALS}  & $\Delta$\textbf{WALS} \\ 
                \hline\hline
                \textbf{IE.Germanic} & en & 1,1,1 &90.5 & 90.3 & 90.4 & -0.1 & 90.4 & 90.6 & +0.1 \\ \hline
                \textbf{IE.Indic}    & ur & -1,-1,1  &18.3 & 35.2 & 34.0 & +15.7 & 35.0 & 33.7 & +15.4 \\
                \textbf{IE.Indic    }& hi & -1,-1,1  &34.3 & 52.4 & 53.4 & +19.1 & 51.3 & 49.1 & +14.8 \\
                \textbf{Dravidian   }& ta & -1,-1,1  &36.1 & 42.8 & 43.4 & +7.3  & 43.1 & 43.0 & +6.9  \\
                \textbf{Turkic      }& tr & -1,-1,1  &31.2 & 35.2 & 37.1 & +5.9  & 35.1 & 36.3 & +5.1  \\
                \textbf{Afro-Asiatic}& ar & 1, 1, -1 &38.5 & 47.3 & 45.3 & +6.8  & 45.8 & 43.7 & +5.2  \\
                \textbf{Afro-Asiatic}& he & 1, 1, -1 &55.7 & 58.8 & 57.6 & +1.9  & 58.3 & 57.6 & +1.9  \\
                \textbf{Austronesian}& id & 1, 1, -1 &49.3 & 53.1 & 52.3 & +3.0  & 52.3 & 51.9 & +2.6  \\
                \textbf{Korean      }& ko & -1,-1,1  &34.0 & 37.1 & 37.2 & +3.2  & 36.3 & 36.4 & +2.4  \\
                \textbf{IE.Celtic   }& cy & 1, 1, -1 &47.3 & 54.2 & 51.7 & +4.4  & 53.8 & 50.0 & +2.7  \\
                \textbf{IE.Romance  }& ca & 1, 1, -1 &73.9 & 74.9 & 73.8 & -0.1  & 74.9 & 74.7 & +0.8  \\
                \textbf{IE.Romance  }& fr & 1, 1, -1 &77.8 & 79.1 & 78.7 & +0.9  & 79.0 & 79.0 & +1.2 \\
                \textbf{Uralic      }& et & 1, -1,1  &65.3 & 65.5 & 65.8 & +0.5  & 65.7 & 66.0 & +0.7  \\
                \textbf{Uralic      }& fi & 1, -1,1  &66.7 & 67.1 & 67.0 & +0.3  & 66.9 & 67.1 & +0.4  \\
                \textbf{IE.Slavic   }& hr & 1, 1, 1  &62.2 & 63.7 & 63.2 & +1.0  & 63.6 & 63.4 & +1.2  \\
                \textbf{IE.Slavic   }& bg & 1, 1, 1  &79.6 & 79.7 & 79.2 & +0.0  & 79.7 & 79.7 & +0.1  \\
                \textbf{IE.Baltic   }& lv & 1, 1, 1  &70.3 & 70.7 & 69.5 & -0.8  & 70.5 & 69.9 & -0.4 \\
                \textbf{IE.Latin    }& la & ?, ?, ?  &47.4 & 48.0 & 45.6 & -1.8  & 48.1 & 47.3 & -0.1 \\
                \textbf{IE.Germanic }& da & 1, 1, 1  &76.6 & 76.6 & 76.5 & -0.1  & 76.6 & 76.6 & +0.0  \\
                \textbf{IE.Germanic }& nl & 0, 1, 1  &67.5 & 67.6 & 67.5 & +0.0  & 67.9 & 67.9 & +0.4  \\
                \hline\hline
                \multicolumn{3}{@{}l@{}|}{\textbf{Average Performance}}& 54.3 & 58.4 & 57.8 & +3.5 & 58.1 & 57.5 & +3.1   \\
            \hline
             \end{tabular}
             \caption{Cross-lingual transfer performances for dependency parsing on 19 languages from 13 different families, with the performance on the source language (English) as a reference. Performances are reported per UAS (we observe similar trends for LAS and details can be found in the appendix Table~\ref{app-table1}).
             We compare the baseline model~\cite{ahmad2018near} with our two algorithms (Lagrangian relaxation and posterior regularization) considering the oracle constraints, and the corpus-statistics constraints compiled from WALS.  Columns $\Delta$WALS denote the improvements bring by leveraging WALS feature as constraints. 
             We also create a Features column to show three WALS features [83A,85A,87A] for each language. The values \{1, -1, 0, ?\} stand for the same as English, opposite to English, no dominant order, and feature missing, respectively. } 
             \label{table1}
             \vspace{-1em}
         \end{table*}

    \subsection{Setup}
    \label{sec:setup}
          \paragraph{Model and Data}
          We train the best performing Att-Graph parser proposed in~\citet{ahmad2018near} on English and transfer it to 19 target languages in UD Tree Bank v2.2 \citep{ud22}.\footnote{We make the selection to prioritize the coverage of language families and low resource languages. The language family information can be found in Table~\ref{table1}.} 
          The model takes words and predicted POS tags\footnote{We use predicted POS tags provided in UD v2.2.} as input, and achieve transfer by leveraging pre-trained multi-lingual FastText \citep{bojanowski2017enriching} embeddings that project the word embeddings from different languages into the same space using an offline transformation method~\citep{smith2017offline,conneau2017word}. 
          The SelfAtt-Graph model uses a Transformer \citep{vaswani2017attention} with relative position embedding as the encoder and a deep biaffine scorer \citep{dozat2017biaffiine} as the decoder.
          We follow the setting in \citet{ahmad2018near} to train and tune only on the source language (English) and directly transfer to all the target languages.  
          We modify their decoder to incorporate constraints with the proposed constrained inference algorithms during the transfer phase without retraining the model. 
          All the hyper-parameters are specified in Appendix Table~\ref{app-hype-table1} together with hyper-parameters for the inference algorithms in Appendix Table~\ref{app-hype-table2}.
          

          \paragraph{Constraints} We consider two types of constraints: 1) instance-level projective constraints for avoiding creating crossing arcs in the dependency trees, 2) corpus-statistics constraints constructed by the process described in Section~\ref{sec:constraintsF}. We consider the following three corpus-statistics constraints: $C1=(NOUN)$, $C2=(NOUN,ADP)$, $C3=(NOUN, ADJ)$; intuitively, $C1$ concerns about the ratio of nouns being on the right of their heads; $C2$ concerns about the ratio of nouns being on the left of adpositions among all noun-adposition arcs; $C3$ concerns about the ratio of nouns being on the left of adjectives among all noun-adjective arcs. 
          
          For binary constraints, $C2$ and $C3$ can be directly compiled from WALS feature $85A$ and $87A$ respectively. We encode ``dominant order'' specified in WALS as the ratio being always greater than $0.75$ (i.e., $r=0.875$ and $\theta=0.125$). If there is no dominant order or the feature is missing, we set $r=0.5$ and $\theta=0.25$. Some WALS features like $82A, 83A$ are also about word order, but we need to specify the arc types to utilize them. For simplicity, we only consider forming constraints from the POS tags in this paper.  
          To estimate the ratio for unary constraint $C1$, we use the WALS features $82A, 83A, 85A, 86A, 87A, 88A, 89A$ that are related to $NOUN$ to form feature vectors, and do regression on languages in the test set except the target language to predict the constraint ratio. 
          The process guarantees the target language remain unseen during the ratio estimation process. 
          The ratios on the regression training languages are estimated by sampling $100$ sentences in the training set per language. 
          
          We also consider an oracle setting where we collect a ``ground-truth'' ratio of each constraint for the target language to estimate an upper bound of our inference algorithms. In the oracle setting, we estimate the ratio on the whole training corpus of the target language and set the margin to $\theta=0.01$.
          

    \subsection{Parsing Performances} 
    \label{sec:result}
        We first compare the performances of the cross-lingual dependency parser with or without constraints. Table~\ref{table1} illustrates the results for the 19 target languages we selected,\footnote{We also run on all languages in \citet{ahmad2018near} for completeness and observe similar trends. The results can be found in Appendix Table~\ref{app-table1}.} along with the performance on the source language (English). 
        The performance on English is not as high as the dependency parsers specialized for English, because to achieve transfer, we have to freeze the pre-trained multi-lingual word embeddings. Yet this parser achieved the best single-source transfer performances according to~\citet{ahmad2018near}.
        
        As is shown in Table~\ref{table1}, the improvements by our constrained inference algorithms are dramatic in a few languages that have very distinct word order features from the source language. For example, the parsing performance of Hindi (hi) improves about 15\% in UAS with WALS features via both Lagrangian relaxation and posterior regularization inference. 
        The improvements are less obvious for languages that are in the same family as English such as Danish(da) and Dutch(nl). This is expected as the corpus linguistic statistics of these languages are similar to English thus the constraints are mostly satisfied with the baseline parser. 
        Comparing Lagrangian relaxation and posterior regularization, we find posterior regularization being more robust and less sensitive to the errors in the corpus-statistics estimation, while Lagrangian relaxation gives a higher improvement on average. Overall, the two proposed constrained inference algorithms improved the transfer performance by 3.5\% and 3.1\% per UAS on average on 19 target languages.
        
        For languages like Finnish (fi) and Estonian (et), the WALS setting works even better than the oracle. We suspect the reason being the large margin we set in the WALS setting. When the estimated corpus-statistics is different from the real ratio in the test set, the large margin relaxes the constraints, thus could result in better performances.  
        
        \paragraph{Discussion.} Despite the major experiments and analysis are conducted using English as the only source language, our approach is general and does not have restriction on the choice of the source language(s). To verify this claim, we run experiments with Hebrew as the source language. Under the oracle setting, Lagrangian relaxation and posterior regularization improve the baseline by 4.4\% and 4.1\%, respectively.
        
        We observed that if we compile WALS features into hard constraints (i.e., set $r$ = 0 or 1), the constraint inference framework only improves performance on half of the languages.  For example, in Estonian (et), the performance \emph{drops} about 3\%. This is because WALS only provides the dominant order. Therefore, treating WALS as hard constraints introduces error to the inference. 
        
        Finally, we assume if we can access to native speakers, the corpus-statistics can be estimated by a few partial annotations of parse trees. In our simulation, using less than 300 arcs, we can achieve the same performance as using the oracle. 
        


        \begin{table}[t]
            \centering
            \begin{tabular}{cccc}
            \hline
                Model & UAS & coverage & $\Delta$ \\
                \hline\hline
                baseline & 54.3 & N/A & N/A\\
                \hline
                +Proj. & 54.6 & N/A & +0.3\\
                \hline
                +Proj.+C1 & 57.0 & 0.24 & +2.4\\
                +Proj.+C2 & 55.7 & 0.08 & +1.1\\
                +Proj.+C3 & 55.0 & 0.07 & +0.4\\
                \hline
                oracle & 58.4 & N/A & +4.1\\
                \hline

            \end{tabular}
            \caption{Ablation study: average UAS of baseline model with different sets of constraints. Proj. represent projective constraints. C1-C3 and oracle are introduced in \ref{sec:setup}. The improvements for projective constraint and oracle are compared to baseline. For the other three constraint sets the improvement is compared to model with projective constraint.}
            \label{table3}
            \vspace{-1em}
        \end{table}
        
    \subsection{Contributions of Individual Constraints}
    \label{sec:ablation}
        We analyze the contribution of each constraint demonstrated in Table~\ref{table3}. Here we use the oracle setting to reduce the noise introduced by corpus-statistics estimation errors. The results are based on Lagrangian relaxation inference. 
        As shown in Table~\ref{table3}, Despite some languages have non-projective dependencies, we observed performance improvements on almost all the languages when the projective constraint is enforced.  All the constraints we formulated have positive contributions to the performance improvements. $C1=(NOUN)$ brings the largest gain probably because its widest coverage. 
        
        Table~\ref{table1} shows that the performance of Hindi improves from 34\% to over 51\% per UAS for both inference algorithms. To better understand where the improvements come from, we conduct an analysis to breakdown the contribution of each individual constraint for Hindi. Table~\ref{table4} shows the results. We can see that since the corpus linguistic statistics between Hindi and English are distinct, the baseline model only achieves low performance. With the constrained inference, especially the postposition constraint (C2), the proposed inference algorithm bring significant improvement.
        
        \begin{table}[t]
        \centering
            \begin{tabular}{c@{\ \  }c@{\ \  }c}
                \hline
                Const. & statistics  & improvement\\
                \hline\hline
                +Proj. & N/A & +0.1\\
                \hline
                C1 & 0.30/0.36/0.94 & +6.9\\
                C2 & 0.00/0.06/1.00 & +11.3\\
                C3 & 0.14/0.27/0.12 & +0.5\\
                \hline
                All & N/A & +18.1\\
                \hline
            \end{tabular}
            \caption{Contribution of individual constraints and their statistics in Hindi. The second column lists the ratios estimated from oracle in English/ baseline in Hindi/ oracle in Hindi, respectively.  The improvement is measured in UAS. The improvement of constraints is computed same as Table \ref{table3}}
            \label{table4}
            \vspace{-1em}
        \end{table}

        To verify the effectiveness of the constraints, we analyze the relation between the performance improvements and corpus statistics ratio gaps between the source and the target languages. 
        To quantify the ratio gap, we weight constraints by their coverage rate and compute the weighted average of the ratio difference between source and target languages. Results show that the performance improvement is highly related to the ratio gap. The Pearson Correlation Coefficient is $\bf 0.938.$ The figure showing the correlation between performance gap (as per UAS) and the corpus statistics ratio gap is in the Appendix Figure~\ref{app-fig:correlation}.

\section{Related Work} \label{sec:related}
\paragraph{Cross-Lingual Transfer for Parsing} 

    Many approaches have been developed to transfer a dependency parser. However, they mainly focus on better capture information from the source language(s). 
    \citet{mcdonald2011multi,guo2016representation,tackstrom2013target,chen2019multi} consider transferring a parser trained on multiple source languages. \citet{agic2017cross,lin2019choosing} selects good source languages by comparing part-of-speech tags sequences. \citet{sogaard2011data, tackstrom2013target} chooses suitable data points from the source language. 
    \citet{pires2019how} uses multilingual BERT to leverage language features from multiple languages. 
    \citet{ahmad2018near} design an order-free model to take out the order features from the source language.
    \citet{xiao2014distributed,guo2015cross} learn an alignment from source words to target words.
    \citet{ponti2018isomorphic} learn an anisomorphism from the source parsing tree to target. 
    \citet{rasooli2019low} reorder the source data before training.
   In contrast, we focus on incorporating linguistic properties in the target languages.

    

\paragraph{Constrained Inference for Parsing} 
    Several previous studies show that adding constraints in inference time improves the performance of models. \citet{grave2015convex} consider incorporating constraints to promote popular types of arcs in an unsupervised setting. \citet{naseem2010using,li2019dependency} train a parser with constraints compiled from the frequency of particular arcs. Compared with the previous work, we focus on cross-lingual transfer with word order constraints.

    Finally, prior studies have noticed that the word order information is significant for parsing and use it as features~\citep{ammar2016many, naseem2012selective, rasooli2017cross, zhang2015hierarchical,dryer2007word}.
    \citet{tackstrom2013target} further propose to decompose these features from models for adapting target languages. \citet{wang2018tacl} use the statistics of surface part-of-speech (POS) tags of target languages to learn the word order. \citet{wang2018synthetic} use POS tags of target languages together with a similar language, and design a stochastic permutation process to synthetic the word order. However, none of them consider using the word order features as constraints. 
    
\paragraph{Incorporating Constraints In NLP Tasks}
    Constraints are widely incorporated in variety of NLP tasks. To name a few, \citet{roth2004linear} propose to formulate constrained inferences in NLP as integer linear programming problems. To solve the intractable structure, \citet{rush2012tutorial} decompose the structure and incorporate constraints on some composite tasks.
    To improve the performance of a model, \citet{chang2011exact,peng2015dual} incorporate constraints on exact decoding tasks and inference tasks on graphical models, and \citet{chang2013tractable,dalvi2015constrained,martins2015transferring} incorporate corpus-level constraints on semi-supervised multilabel classification and coreference resolution.
    \citet{jieyu2017men} incorporate corpus-level constraints to avoid amplifying gender bias on visual semantic role labeling and multilabel classification. 
    In contrast to previous work, we incorporate corpus-level constraints to facilitate dependency parser in the cross-lingual transfer setting.
        
\section{Conclusion}
    We propose to leverage corpus-linguistic statistics to guide the inference of cross-lingual dependency parsing. 
    We compile these statistics into corpus-statistic constraints and design two inference algorithms on top of a graph-based parser based on Lagrangian relaxation and posterior regularization. 
    Experiments on 19 languages show that our approach improves the performance of the cross-lingual parser substantially. 
    In the future, we plan to study the design and incorporation of fine-grained constraints considering multipule languages for cross-lingual transfer. 
    We also plan to adapt this constrained inference framework to other cross-lingual structured prediction problems, such as semantic role labeling. 

\paragraph{Acknowledgement} This work was supported in part by National Science Foundation Grant IIS-1760523 and an  NIH R01 (LM012592). We thank anonymous reviewers and members of the UCLA-NLP lab for their feedback.

\bibliography{emnlp-ijcnlp-2019}

\begin{thebibliography}{53}
\expandafter\ifx\csname natexlab\endcsname\relax\def\natexlab#1{#1}\fi

\bibitem[{Agi{\'c}(2017)}]{agic2017cross}
{\v{Z}}eljko Agi{\'c}. 2017.
\newblock Cross-lingual parser selection for low-resource languages.
\newblock In \emph{Proceedings of the NoDaLiDa Workshop on Universal
  Dependencies, UDW@NoDaLiDa 2017, Gothenburg, Sweden, May 22, 2017}, pages
  1--10. Association for Computational Linguistics.

\bibitem[{Agi{\'c} et~al.(2014)Agi{\'c}, Tiedemann, Dobrovoljc, Krek, Merkler,
  and Mo{\v{z}}e}]{agic2014cross}
{\v{Z}}eljko Agi{\'c}, J{\"o}rg Tiedemann, Kaja Dobrovoljc, Simon Krek,
  Danijela Merkler, and Sara Mo{\v{z}}e. 2014.
\newblock Cross-lingual dependency parsing of related languages with rich
  morphosyntactic tagsets.
\newblock In \emph{EMNLP 2014 Workshop on Language Technology for Closely
  Related Languages and Language Variants}.

\bibitem[{Ahmad et~al.(2019)Ahmad, Zhang, Ma, Hovy, Chang, and
  Peng}]{ahmad2018near}
Wasi~Uddin Ahmad, Zhisong Zhang, Xuezhe Ma, Eduard Hovy, Kai-Wei Chang, and
  Nanyun Peng. 2019.
\newblock On difficulties of cross-lingual transfer with order differences: A
  case study on dependency parsing.
\newblock In \emph{Proceedings of the 2019 Conference of the North American
  Chapter of the Association for Computational Linguistics}.

\bibitem[{Ammar et~al.(2016)Ammar, Mulcaire, Ballesteros, Dyer, and
  Smith}]{ammar2016many}
Waleed Ammar, George Mulcaire, Miguel Ballesteros, Chris Dyer, and Noah~A
  Smith. 2016.
\newblock Many languages, one parser.
\newblock \emph{Transactions of the Association for Computational Linguistics},
  4:431--444.

\bibitem[{Bojanowski et~al.(2017)Bojanowski, Grave, Joulin, and
  Mikolov}]{bojanowski2017enriching}
Piotr Bojanowski, Edouard Grave, Armand Joulin, and Tomas Mikolov. 2017.
\newblock Enriching word vectors with subword information.
\newblock \emph{Transactions of the Association for Computational Linguistics},
  5:135--146.

\bibitem[{Chang et~al.(2013)Chang, Sundararajan, and
  Keerthi}]{chang2013tractable}
Kai{-}Wei Chang, S.~Sundararajan, and S.~Sathiya Keerthi. 2013.
\newblock Tractable semi-supervised learning of complex structured prediction
  models.
\newblock In \emph{Machine Learning and Knowledge Discovery in Databases -
  European Conference, {ECML} {PKDD} 2013, Prague, Czech Republic, September
  23-27, 2013, Proceedings, Part {III}}, volume 8190 of \emph{Lecture Notes in
  Computer Science}, pages 176--191. Springer.

\bibitem[{Chang and Collins(2011)}]{chang2011exact}
Yin{-}Wen Chang and Michael Collins. 2011.
\newblock Exact decoding of phrase-based translation models through lagrangian
  relaxation.
\newblock In \emph{Proceedings of the 2011 Conference on Empirical Methods in
  Natural Language Processing, {EMNLP} 2011, 27-31 July 2011, John McIntyre
  Conference Centre, Edinburgh, UK, {A} meeting of SIGDAT, a Special Interest
  Group of the {ACL}}, pages 26--37. {ACL}.

\bibitem[{Chen et~al.(2019)Chen, Awadallah, Hassan, Wang, and
  Cardie}]{chen2019multi}
Xilun Chen, Ahmed~Hassan Awadallah, Hany Hassan, Wei Wang, and Claire Cardie.
  2019.
\newblock Multi-source cross-lingual model transfer: Learning what to share.
\newblock In \emph{Proceedings of the 57th Conference of the Association for
  Computational Linguistics, {ACL} 2019, Florence, Italy, July 28- August 2,
  2019, Volume 1: Long Papers}, pages 3098--3112. Association for Computational
  Linguistics.

\bibitem[{Chu and Liu(1965)}]{chu-liu-1965}
Y.~J. Chu and T.~H. Liu. 1965.
\newblock {On the shortest arborescence of a directed graph}.
\newblock \emph{Science Sinica}, 14.

\bibitem[{Conneau et~al.(2018)Conneau, Lample, Ranzato, Denoyer, and
  J{\'e}gou}]{conneau2017word}
Alexis Conneau, Guillaume Lample, Marc'Aurelio Ranzato, Ludovic Denoyer, and
  Herv{\'e} J{\'e}gou. 2018.
\newblock Word translation without parallel data.
\newblock \emph{Internation Conference on Learning Representations}.

\bibitem[{Dalvi(2015)}]{dalvi2015constrained}
Bhavana~Bharat Dalvi. 2015.
\newblock \emph{Constrained Semi-supervised Learning in the Presence of
  Unanticipated Classes}.
\newblock Ph.D. thesis, Google Research.

\bibitem[{Dozat and Manning(2017)}]{dozat2017biaffiine}
Timothy Dozat and Christopher~D Manning. 2017.
\newblock Deep biaffine attention for neural dependency parsing.
\newblock \emph{Internation Conference on Learning Representations}.

\bibitem[{Dryer(2007)}]{dryer2007word}
Matthew~S Dryer. 2007.
\newblock Word order.
\newblock \emph{Language typology and syntactic description}, 1:61--131.

\bibitem[{Dryer and Haspelmath(2013)}]{wals}
Matthew~S. Dryer and Martin Haspelmath, editors. 2013.
\newblock \emph{WALS Online}.
\newblock Max Planck Institute for Evolutionary Anthropology, Leipzig.

\bibitem[{Ganchev et~al.(2010)Ganchev, Gillenwater, Taskar
  et~al.}]{ganchev2010posterior}
Kuzman Ganchev, Jennifer Gillenwater, Ben Taskar, et~al. 2010.
\newblock Posterior regularization for structured latent variable models.
\newblock \emph{Journal of Machine Learning Research}, 11(Jul):2001--2049.

\bibitem[{Grave and Elhadad(2015)}]{grave2015convex}
Edouard Grave and No{\'e}mie Elhadad. 2015.
\newblock A convex and feature-rich discriminative approach to dependency
  grammar induction.
\newblock In \emph{Proceedings of the 53rd Annual Meeting of the Association
  for Computational Linguistics and the 7th International Joint Conference on
  Natural Language Processing (Volume 1: Long Papers)}, volume~1, pages
  1375--1384.

\bibitem[{Guo et~al.(2015)Guo, Che, Yarowsky, Wang, and Liu}]{guo2015cross}
Jiang Guo, Wanxiang Che, David Yarowsky, Haifeng Wang, and Ting Liu. 2015.
\newblock Cross-lingual dependency parsing based on distributed
  representations.
\newblock In \emph{Proceedings of the 53rd Annual Meeting of the Association
  for Computational Linguistics and the 7th International Joint Conference on
  Natural Language Processing (Volume 1: Long Papers)}, volume~1, pages
  1234--1244.

\bibitem[{Guo et~al.(2016)Guo, Che, Yarowsky, Wang, and
  Liu}]{guo2016representation}
Jiang Guo, Wanxiang Che, David Yarowsky, Haifeng Wang, and Ting Liu. 2016.
\newblock A representation learning framework for multi-source transfer
  parsing.
\newblock In \emph{Proceedings of the Thirtieth AAAI Conference on Artificial
  Intelligence}, AAAI'16, pages 2734--2740.

\bibitem[{Joty et~al.(2017)Joty, Nakov, M{\`a}rquez, and
  Jaradat}]{joty2017cross}
Shafiq Joty, Preslav Nakov, Llu{\'\i}s M{\`a}rquez, and Israa Jaradat. 2017.
\newblock Cross-language learning with adversarial neural networks.
\newblock In \emph{Proceedings of the 21st Conference on Computational Natural
  Language Learning (CoNLL 2017)}, pages 226--237.

\bibitem[{Kingma and Ba(2015)}]{kingma2014adam}
Diederik~P. Kingma and Jimmy Ba. 2015.
\newblock Adam: {A} method for stochastic optimization.
\newblock In \emph{3rd International Conference on Learning Representations,
  {ICLR} 2015, San Diego, CA, USA, May 7-9, 2015, Conference Track
  Proceedings}.

\bibitem[{Kiperwasser and Goldberg(2016)}]{kiperwasser2016simple}
Eliyahu Kiperwasser and Yoav Goldberg. 2016.
\newblock Simple and accurate dependency parsing using bidirectional lstm
  feature representations.
\newblock \emph{Transactions of the Association for Computational Linguistics},
  4:313--327.

\bibitem[{Kuncoro et~al.(2016)Kuncoro, Ballesteros, Kong, Dyer, and
  Smith}]{kuncoro2016distilling}
Adhiguna Kuncoro, Miguel Ballesteros, Lingpeng Kong, Chris Dyer, and Noah~A
  Smith. 2016.
\newblock Distilling an ensemble of greedy dependency parsers into one mst
  parser.
\newblock In \emph{Proceedings of the 2016 Conference on Empirical Methods in
  Natural Language Processing}, pages 1744--1753.

\bibitem[{Li et~al.(2019)Li, Cheng, Liu, and Keller}]{li2019dependency}
Bowen Li, Jianpeng Cheng, Yang Liu, and Frank Keller. 2019.
\newblock Dependency grammar induction with a neural variational
  transition-based parser.
\newblock In \emph{Proceedings of the AAAI Conference on Artificial
  Intelligence}, volume~33, pages 6658--6665.

\bibitem[{Lin et~al.(2019)Lin, Chen, Lee, Li, Zhang, Xia, Rijhwani, He, Zhang,
  Ma, Anastasopoulos, Littell, and Neubig}]{lin2019choosing}
Yu{-}Hsiang Lin, Chian{-}Yu Chen, Jean Lee, Zirui Li, Yuyan Zhang, Mengzhou
  Xia, Shruti Rijhwani, Junxian He, Zhisong Zhang, Xuezhe Ma, Antonios
  Anastasopoulos, Patrick Littell, and Graham Neubig. 2019.
\newblock Choosing transfer languages for cross-lingual learning.
\newblock In \emph{Proceedings of the 57th Conference of the Association for
  Computational Linguistics, {ACL} 2019, Florence, Italy, July 28- August 2,
  2019, Volume 1: Long Papers}, pages 3125--3135. Association for Computational
  Linguistics.

\bibitem[{Martins(2015)}]{martins2015transferring}
Andr{\'{e}} F.~T. Martins. 2015.
\newblock Transferring coreference resolvers with posterior regularization.
\newblock In \emph{Proceedings of the 53rd Annual Meeting of the Association
  for Computational Linguistics and the 7th International Joint Conference on
  Natural Language Processing of the Asian Federation of Natural Language
  Processing, ACL 2015, July 26-31, 2015, Beijing, China, Volume 1: Long
  Papers}, pages 1427--1437. The Association for Computer Linguistics.

\bibitem[{McDonald et~al.(2005)McDonald, Crammer, and Pereira}]{McDonald:2005}
Ryan McDonald, Koby Crammer, and Fernando Pereira. 2005.
\newblock Online large-margin training of dependency parsers.
\newblock In \emph{Proceedings of ACL-2005}, pages 91--98, Ann Arbor, Michigan,
  USA.

\bibitem[{McDonald et~al.(2013)McDonald, Nivre, Quirmbach-Brundage, Goldberg,
  Das, Ganchev, Hall, Petrov, Zhang, T{\"a}ckstr{\"o}m
  et~al.}]{mcdonald2013universal}
Ryan McDonald, Joakim Nivre, Yvonne Quirmbach-Brundage, Yoav Goldberg, Dipanjan
  Das, Kuzman Ganchev, Keith Hall, Slav Petrov, Hao Zhang, Oscar
  T{\"a}ckstr{\"o}m, et~al. 2013.
\newblock Universal dependency annotation for multilingual parsing.
\newblock In \emph{Proceedings of the 51st Annual Meeting of the Association
  for Computational Linguistics (Volume 2: Short Papers)}, volume~2, pages
  92--97.

\bibitem[{McDonald et~al.(2011)McDonald, Petrov, and Hall}]{mcdonald2011multi}
Ryan McDonald, Slav Petrov, and Keith Hall. 2011.
\newblock Multi-source transfer of delexicalized dependency parsers.
\newblock In \emph{Proceedings of the conference on empirical methods in
  natural language processing}, pages 62--72. Association for Computational
  Linguistics.

\bibitem[{Naseem et~al.(2012)Naseem, Barzilay, and
  Globerson}]{naseem2012selective}
Tahira Naseem, Regina Barzilay, and Amir Globerson. 2012.
\newblock Selective sharing for multilingual dependency parsing.
\newblock In \emph{Proceedings of the 50th Annual Meeting of the Association
  for Computational Linguistics: Long Papers-Volume 1}, pages 629--637.
  Association for Computational Linguistics.

\bibitem[{Naseem et~al.(2010)Naseem, Chen, Barzilay, and
  Johnson}]{naseem2010using}
Tahira Naseem, Harr Chen, Regina Barzilay, and Mark Johnson. 2010.
\newblock Using universal linguistic knowledge to guide grammar induction.
\newblock In \emph{Proceedings of the 2010 Conference on Empirical Methods in
  Natural Language Processing}, pages 1234--1244. Association for Computational
  Linguistics.

\bibitem[{Nivre et~al.(2018)Nivre, Abrams, Agi{\'c}, and et~al.}]{ud22}
Joakim Nivre, Mitchell Abrams, {\v Z}eljko Agi{\'c}, and et~al. 2018.
\newblock Universal dependencies 2.2.
\newblock {LINDAT}/{CLARIN} digital library at the Institute of Formal and
  Applied Linguistics ({{\'U}FAL}), Faculty of Mathematics and Physics, Charles
  University.

\bibitem[{{\"O}stling(2015)}]{ostling2015word}
Robert {\"O}stling. 2015.
\newblock Word order typology through multilingual word alignment.
\newblock In \emph{Proceedings of the 53rd Annual Meeting of the Association
  for Computational Linguistics and the 7th International Joint Conference on
  Natural Language Processing (Volume 2: Short Papers)}, volume~2, pages
  205--211.

\bibitem[{Peng et~al.(2015)Peng, Cotterell, and Eisner}]{peng2015dual}
Nanyun Peng, Ryan Cotterell, and Jason Eisner. 2015.
\newblock Dual decomposition inference for graphical models over strings.
\newblock In \emph{Proceedings of the 2015 Conference on Empirical Methods in
  Natural Language Processing, {EMNLP} 2015, Lisbon, Portugal, September 17-21,
  2015}, pages 917--927. The Association for Computational Linguistics.

\bibitem[{Pires et~al.(2019)Pires, Schlinger, and Garrette}]{pires2019how}
Telmo Pires, Eva Schlinger, and Dan Garrette. 2019.
\newblock How multilingual is multilingual bert?
\newblock In \emph{Proceedings of the 57th Conference of the Association for
  Computational Linguistics, {ACL} 2019, Florence, Italy, July 28- August 2,
  2019, Volume 1: Long Papers}, pages 4996--5001. Association for Computational
  Linguistics.

\bibitem[{Ponti et~al.(2018)Ponti, Reichart, Korhonen, and
  Vuli{\'c}}]{ponti2018isomorphic}
Edoardo~Maria Ponti, Roi Reichart, Anna Korhonen, and Ivan Vuli{\'c}. 2018.
\newblock Isomorphic transfer of syntactic structures in cross-lingual {NLP}.
\newblock In \emph{Proceedings of the 56th Annual Meeting of the Association
  for Computational Linguistics, {ACL} 2018, Melbourne, Australia, July 15-20,
  2018, Volume 1: Long Papers}, pages 1531--1542. Association for Computational
  Linguistics.

\bibitem[{Rasooli and Collins(2017)}]{rasooli2017cross}
Mohammad~Sadegh Rasooli and Michael Collins. 2017.
\newblock Cross-lingual syntactic transfer with limited resources.
\newblock \emph{Transactions of the Association for Computational Linguistics},
  5:279--293.

\bibitem[{Rasooli and Collins(2019)}]{rasooli2019low}
Mohammad~Sadegh Rasooli and Michael Collins. 2019.
\newblock Low-resource syntactic transfer with unsupervised source reordering.
\newblock In \emph{Proceedings of the 2019 Conference of the North American
  Chapter of the Association for Computational Linguistics: Human Language
  Technologies, {NAACL-HLT} 2019, Minneapolis, MN, USA, June 2-7, 2019, Volume
  1 (Long and Short Papers)}, pages 3845--3856. Association for Computational
  Linguistics.

\bibitem[{Roth and Yih(2004)}]{roth2004linear}
Dan Roth and Wen-tau Yih. 2004.
\newblock A linear programming formulation for global inference in natural
  language tasks.
\newblock Technical report, ILLINOIS UNIV AT URBANA-CHAMPAIGN DEPT OF COMPUTER
  SCIENCE.

\bibitem[{Rush and Collins(2011)}]{rush2011exact}
Alexander~M Rush and Michael Collins. 2011.
\newblock Exact decoding of syntactic translation models through {Lagrangian}
  relaxation.
\newblock In \emph{Proceedings of the 49th Annual Meeting of the Association
  for Computational Linguistics: Human Language Technologies}, pages 72--82.

\bibitem[{Rush and Collins(2012)}]{rush2012tutorial}
Alexander~M Rush and MJ~Collins. 2012.
\newblock A tutorial on dual decomposition and lagrangian relaxation for
  inference in natural language processing.
\newblock \emph{Journal of Artificial Intelligence Research}, 45:305--362.

\bibitem[{Smith et~al.(2017)Smith, Turban, Hamblin, and
  Hammerla}]{smith2017offline}
Samuel~L Smith, David~HP Turban, Steven Hamblin, and Nils~Y Hammerla. 2017.
\newblock Offline bilingual word vectors, orthogonal transformations and the
  inverted softmax.
\newblock \emph{Internation Conference on Learning Representations}.

\bibitem[{S{\o}gaard(2011)}]{sogaard2011data}
Anders S{\o}gaard. 2011.
\newblock Data point selection for cross-language adaptation of dependency
  parsers.
\newblock In \emph{Proceedings of the 49th Annual Meeting of the Association
  for Computational Linguistics: Human Language Technologies: short
  papers-Volume 2}, pages 682--686. Association for Computational Linguistics.

\bibitem[{T{\"a}ckstr{\"o}m et~al.(2013)T{\"a}ckstr{\"o}m, McDonald, and
  Nivre}]{tackstrom2013target}
Oscar T{\"a}ckstr{\"o}m, Ryan McDonald, and Joakim Nivre. 2013.
\newblock Target language adaptation of discriminative transfer parsers.
\newblock In \emph{Proceedings of the 2013 Conference of the North American
  Chapter of the Association for Computational Linguistics: Human Language
  Technologies}, pages 1061--1071. Association for Computational Linguistics.

\bibitem[{Tiedemann(2015)}]{tiedemann2015cross}
J{\"o}rg Tiedemann. 2015.
\newblock Cross-lingual dependency parsing with universal dependencies and
  predicted pos labels.
\newblock In \emph{Proceedings of the Third International Conference on
  Dependency Linguistics (Depling 2015)}, pages 340--349.

\bibitem[{Vaswani et~al.(2017)Vaswani, Shazeer, Parmar, Uszkoreit, Jones,
  Gomez, Kaiser, and Polosukhin}]{vaswani2017attention}
Ashish Vaswani, Noam Shazeer, Niki Parmar, Jakob Uszkoreit, Llion Jones,
  Aidan~N Gomez, {\L}ukasz Kaiser, and Illia Polosukhin. 2017.
\newblock Attention is all you need.
\newblock In \emph{Advances in Neural Information Processing Systems}, pages
  5998--6008.

\bibitem[{Wang and Eisner(2018{\natexlab{a}})}]{wang2018tacl}
Dingquan Wang and Jason Eisner. 2018{\natexlab{a}}.
\newblock Surface statistics of an unknown language indicate how to parse it.
\newblock \emph{Transactions of the Association for Computational Linguistics
  (TACL)}.

\bibitem[{Wang and Eisner(2018{\natexlab{b}})}]{wang2018synthetic}
Dingquan Wang and Jason Eisner. 2018{\natexlab{b}}.
\newblock Synthetic data made to order: The case of parsing.
\newblock In \emph{Proceedings of the 2018 Conference on Empirical Methods in
  Natural Language Processing}, pages 1325--1337.

\bibitem[{Wang and Chang(2016)}]{wang2016graph}
Wenhui Wang and Baobao Chang. 2016.
\newblock Graph-based dependency parsing with bidirectional lstm.
\newblock In \emph{Proceedings of the 54th Annual Meeting of the Association
  for Computational Linguistics (Volume 1: Long Papers)}, volume~1, pages
  2306--2315.

\bibitem[{Xiao and Guo(2014)}]{xiao2014distributed}
Min Xiao and Yuhong Guo. 2014.
\newblock Distributed word representation learning for cross-lingual dependency
  parsing.
\newblock In \emph{Proceedings of the Eighteenth Conference on Computational
  Natural Language Learning}, pages 119--129.

\bibitem[{Xie et~al.(2018)Xie, Yang, Neubig, Smith, and
  Carbonell}]{xie2018neural}
Jiateng Xie, Zhilin Yang, Graham Neubig, Noah~A. Smith, and Jaime Carbonell.
  2018.
\newblock Neural cross-lingual named entity recognition with minimal resources.
\newblock In \emph{Proceedings of the 2018 Conference on Empirical Methods in
  Natural Language Processing}, pages 369--379. Association for Computational
  Linguistics.

\bibitem[{Zeman and Resnik(2008)}]{zeman2008cross}
Daniel Zeman and Philip Resnik. 2008.
\newblock Cross-language parser adaptation between related languages.
\newblock In \emph{Proceedings of the IJCNLP-08 Workshop on NLP for Less
  Privileged Languages}.

\bibitem[{Zhang and Barzilay(2015)}]{zhang2015hierarchical}
Yuan Zhang and Regina Barzilay. 2015.
\newblock Hierarchical low-rank tensors for multilingual transfer parsing.
\newblock Association for Computational Linguistics.

\bibitem[{Zhao et~al.(2017)Zhao, Wang, Yatskar, Ordonez, and
  Chang}]{jieyu2017men}
Jieyu Zhao, Tianlu Wang, Mark Yatskar, Vicente Ordonez, and Kai{-}Wei Chang.
  2017.
\newblock Men also like shopping: Reducing gender bias amplification using
  corpus-level constraints.
\newblock In \emph{Proceedings of the 2017 Conference on Empirical Methods in
  Natural Language Processing, {EMNLP} 2017, Copenhagen, Denmark, September
  9-11, 2017}, pages 2979--2989. Association for Computational Linguistics.

\end{thebibliography}
\bibliographystyle{acl_natbib}

\clearpage

\appendix
\section{Details in Posterior Regularization}
    \label{app-sec:details}
    \paragraph{Expectation rewriting}
        \begin{small}
        \begin{eqnarray*}
            R(C,q) &\leq& r \\
            \frac{\sum\limits_k\sum\limits_{(i,j):(k,i,j)\in C^+}q_k(i,j)}{\sum\limits_k\sum\limits_{(i,j):(k,i,j)\in C^+ \cup C^-}q_k(i,j)} &\leq& r \\
            (1-r)\sum_{(k,i,j)\in C^+}q_k(i,j)-r\sum_{(k,i,j)\in C^-}q_k(i,j) &\leq& 0 \\
            \sum_{(k,i,j)}q_k(i,j)\phi(k,i,j) &\leq& 0 \\
            \mathbb{E}_{\mathbf{y}\sim q}[\phi(\mathbf{y})] &\leq& 0 \\
        \end{eqnarray*}
        \end{small}
        Let $\phi'(k,i,j)=-\phi(k,i,j),$ we have
        
        \begin{small}
        \begin{eqnarray*}
            R(C,q) &\geq& r \\
            \frac{\sum\limits_k\sum\limits_{(i,j):(k,i,j)\in C^+}q_k(i,j)}{\sum\limits_k\sum\limits_{(i,j):(k,i,j)\in C^+ \cup C^-}q_k(i,j)} &\geq& r \\
            (1-r)\sum_{(k,i,j)\in C^+}q_k(i,j)-r\sum_{(k,i,j)\in C^-}q_k(i,j) &\geq& 0 \\
            \sum_{(k,i,j)}q_k(i,j)\phi(k,i,j) &\geq& 0 \\
            \sum_{(k,i,j)}q_k(i,j)\phi'(k,i,j) &\leq& 0 \\
            \mathbb{E}_{\mathbf{y}\sim q}[\phi'(\mathbf{y})] &\leq& 0 \\
        \end{eqnarray*}
        \end{small}
    
    \paragraph{Dual form solving}
    This dual form of the optimization problem is given in \citet{ganchev2010posterior} as
    $$q^*(\mathbf{y})=\frac{p_\theta(\mathbf{y}|\mathbf{w})\exp(-\lambda^*\cdot\phi(\mathbf{y}))}{Z(\lambda^*)},$$
    $$\lambda^*=\arg\max\limits_{\lambda\geq 0}-\log Z(\lambda),$$
    $$Z(\lambda)=\sum\limits_{\mathbf{y}'}p_\theta(\mathbf{y}'|\mathbf{w})\exp(-\lambda^*\cdot\phi(\mathbf{y}')).$$
    
    Noting that both the feature function $\phi$ and model $p_\theta$ can be easily factorized to arc-level, we can set
    $$q^*_k(i,j)=p_\theta(y_k(i,j)|\mathbf{w}_k)\exp(-\lambda^*\cdot \phi(k,i,j))$$
    as our model on target language.
    
    The dual form is not easy to solve since the number of possible $\mathbf{y}'$ can be exponentially large. We need to do factorization to make it tractable.
    
    \begin{small}
        \begin{eqnarray*}
            & & Z(\lambda)\\
            &=& \sum\limits_{\mathbf{y}'}p_\theta(\mathbf{y}'|\mathbf{w})\exp(-\lambda\cdot\phi(\mathbf{w},\mathbf{y}'))\\
            &=& \sum\limits_{\mathbf{y}'}\prod\limits_{k}p_\theta(\mathbf{y}_k|\mathbf{w}_k)\exp(-\lambda\cdot\phi(\mathbf{w}_k,\mathbf{y}_k))\\
            &=& \sum\limits_{\mathbf{y}'}\prod\limits_{k}\prod_{(i,j):\mathbf{y}_k(i,j)=1}p_\theta(y_k(i,j)|\mathbf{w}_k)\exp(-\lambda\cdot\phi(y_k(i,j))\\
            &=& \sum\limits_{\mathbf{y}'}\prod\limits_{k}\prod_{(i,j):\mathbf{y}_k(i,j)=1}q^*_k(i,j)\\
            &=& \prod\limits_{k\in[N]}\prod_{i\in[L_k]}\sum\limits_{j\in[L_k]}q^*_k(i,j)\\
            \\
            & & \log Z(\lambda)\\
            &=& \sum\limits_{k\in[N]}\sum_{i\in[L_k]}\log\sum\limits_{j\in[L_k]}q^*_k(i,j)\\
            & & \frac{\partial \log Z(\lambda)}{\partial\lambda}\\
            &=&\sum\limits_{k\in[N]}\sum_{i\in[L_k]}\log\frac{\sum\limits_{j\in[L_k]} -\phi(k,i,j)q^*_k(i,j)}{\sum\limits_{j\in[L_k]}q^*_k(i,j)}
        \end{eqnarray*}
    \end{small}
    
    The Hessian matrix of $Z(\lambda)$ is given by
    $$H(Z(\lambda))=\sum\limits_{\mathbf{y}'}p_\theta(\mathbf{y}'|\mathbf{w})\exp(-\lambda\cdot\phi)[\phi^T\phi].$$
    Noting that $\phi^T\phi$ is positive semi-definite, and $p_\theta(\mathbf{y}'|\mathbf{w})\exp(-\lambda\cdot\phi)\geq 0$, so $H(Z(\lambda))$ is also positive semi-definite, which means $Z(\lambda)$ is convex, and $\log Z(\lambda)$ is quasi-convex. We can sample $b$ instances from dataset and compute the gradient to estimate the full gradient, and apply stochastic gradient descent to get the optimal $\lambda^*$.
    We can verify that $Z(\lambda)$ is convex, and $\log Z(\lambda)$ is quasi-convex. We can sample $b$ instances from dataset and compute the gradient to estimate the full gradient, and apply stochastic gradient descent or Adam to get the optimal $\lambda^*$.

\section{Hyper-parameters}
    \label{app-sec:hyper}
    \paragraph{Model}
       We follow the hyper-parameters used in \citet{ahmad2018near} shown in table \ref{app-hype-table1}.
       \begin{table}[t]
            \centering
            \begin{tabular}{ll}
                \hline
                Hyper-parameter & Value \\
                \hline
                Input word dimension & 300\\
                Input pos dimension & 50\\
                Encoder layer & 6\\
                Encoder $d_{model}$ & 350\\
                Encoder $d_{ff}$ & 512 \\
                Arc MLP size & 512\\
                Label MLP size & 128\\
                Training dropout & 0.2\\
                Optimizer & Adam\\
                Learning rate & 0.0001\\
                Batch size & 80\\
                \hline
            \end{tabular}
            \caption{Hyper-parameters in our model.}
            \label{app-hype-table1}
        \end{table}
    \paragraph{Inference}
       The hyper-parameters in inference algorithms, Lagrangian relaxation and posterior regularization, are shown in table \ref{app-hype-table2}.
       \begin{table}[t]
            \centering
            \begin{tabular}{lll}
                \hline
                Hyper-parameter & Lagrangian & PR \\
                \hline
                Initial learning rate & 50 & 1 \\
                Learning rate decay & 0.9 & 0.98 \\
                Maximal iteration & 60 & 100 \\
                Batch size & full batch & 128\\
                \hline
            \end{tabular}
            \caption{Hyper-parameters in our model.}
            \label{app-hype-table2}
        \end{table}

\section{Language Family}
\label{app-sec:family}
    The languages we selected and their language families are shown in Table \ref{app-tab:langs}
    \begin{table}[t!]
        \centering
        \small
        \begin{tabular}{p{1.8cm} |p{4.75cm}}
            \hline
            Language Families & Languages \\
            \hline
            Afro-Asiatic & Arabic (ar), Hebrew (he)\\
            \hline
            Austronesian & Indonesian (id)\\
            \hline
            Dravidian & Tamil (ta)\\
            \hline
            Turkic & Turkish (tr)\\
            \hline
            IE.Celtic & Welsh (cy)\\
            \hline
            IE.Baltic & Latvian (lv)\\
            \hline
            IE.Germanic & Danish (da), Dutch (nl), English (en)\\
            \hline
            IE.Indic & Hindi (hi), Urdu (ur)\\
            \hline
            IE.Latin & Latin (la)\\
            \hline
            IE.Romance & Catalan (ca), French (fr) \\
            \hline
            IE.Slavic & Bulgarian (bg), Croatian (hr)\\
            \hline
            Korean & Korean (ko)\\
            \hline
            Uralic & Estonian (et), Finnish (fi)\\
            \hline
        \end{tabular}
    \caption{
    \label{app-tab:langs}
    The selected languages for experiments from UD v2.2 \cite{ud22}).
    }
    \vspace{-1em}
    \end{table}

\section{Entire experiment results}
\label{app-sec:entire}
    The entire experiment results are shown in Table \ref{app-table1}
    \begin{table*}[t!]
        \begin{tabular}{@{}l@{\ \ } |@{}c@{}| @{ }c@{ } | @{ }c@{ } | c@{\  \ }c@{\ \ }c | c@{\ \ }c@{\ \ }c@{}}
            \hline
            \multirow{2}{*}{\textbf{Family}} & \multirow{2}{*}{\textbf{Lang.}}  & \multirow{2}{*}{\textbf{Features}} & \multirow{2}{*}{\textbf{Baseline}}  & \multicolumn{3}{c|}{\textbf{Lagrangian Relaxation}} & \multicolumn{3}{c}{\textbf{Posterior Regularization}}\\
            \cline{5-10}
            & & & & \textbf{Oracle} & \textbf{WALS}  & $\Delta$\textbf{WALS} & \textbf{Oracle} & \textbf{WALS}  & $\Delta$\textbf{WALS} \\ 
        \hline\hline
        \textbf{IE.Indic    } & ur & -1,-1,1  & 18.3 & 35.2 & 34.0 & +15.7 & 35.0 & 33.7 & +15.4 \\
        \textbf{IE.Indic    } & hi & -1,-1,1  & 34.3 & 52.4 & 53.4 & +19.1 & 51.3 & 49.1 & +14.8 \\
        \textbf{Dravidian   } & ta & -1,-1,1  & 36.1 & 42.8 & 43.4 & +7.3  & 43.1 & 43.0 & +6.9  \\
        \textbf{Turkic      } & tr & -1,-1,1  & 31.2 & 35.2 & 37.1 & +5.9  & 35.1 & 36.3 & +5.1  \\
        \textbf{Afro-Asiatic} & ar & 1, 1,-1  & 38.5 & 47.3 & 45.3 & +6.8  & 45.8 & 43.7 & +5.2  \\
        \textbf{Afro-Asiatic} & he & 1, 1, 1  & 55.7 & 58.8 & 57.6 & +1.9  & 58.3 & 57.6 & +1.9  \\
        \textbf{Austronesian} & id & 1, 1, 1  & 49.3 & 53.1 & 52.3 & +3.0  & 52.3 & 51.9 & +2.6  \\
        \textbf{Korean      } & ko & -1,-1,1  & 34.0 & 37.1 & 37.2 & +3.2  & 36.3 & 36.4 & +2.4  \\
        \textbf{IE.Celtic   } & cy & 1, 1,-1  & 47.3 & 54.2 & 51.7 & +4.4  & 53.8 & 50.0 & +2.7  \\
        \textbf{IE.Slavic   } & hr & 1, 1, 1  & 62.2 & 63.7 & 63.2 & +1.0  & 63.6 & 63.4 & +1.2  \\
        \textbf{IE.Slavic   } & bg & 1, 1, 1  & 79.6 & 79.7 & 79.2 & +0.0  & 79.7 & 79.7 & +0.1  \\
        \textbf{IE.Slavic   } & cs & 1, 1, 1  & 63.0 & 63.9 & 64.0 & +1.0  & 63.8 & 63.6 & +0.6  \\
        \textbf{IE.Slavic   } & pl & 1, 1, 1  & 74.6 & 74.8 & 73.6 & -1.0  & 75.0 & 74.8 & +0.2  \\
        \textbf{IE.Slavic   } & ru & 1, 1, 1  & 60.6 & 61.6 & 61.2 & +0.6  & 61.4 & 61.4 & +0.8  \\
        \textbf{IE.Slavic   } & sk & ?, ?, ? & 66.8 & 66.3 & 67.9 & +1.1  & 67.0 & 66.8 & +0.0  \\
        \textbf{IE.Slavic   } & sl & 1, 1, 1  & 67.8 & 67.9 & 67.9 & +0.1  & 67.9 & 67.9 & +0.1  \\
        \textbf{IE.Slavic   } & uk & 1, 1, 1  & 59.9 & 62.1 & 60.9 & +1.0  & 61.9 & 61.1 & +0.2  \\
        \textbf{IE.Romance  } & ca & 1, 1,-1  & 73.9 & 74.9 & 73.8 & -0.1  & 74.9 & 74.7 & +0.8  \\
        \textbf{IE.Romance  } & fr & 1, 1,-1  & 77.8 & 79.1 & 78.7 & +0.9  & 79.0 & 79.0 & +1.2  \\
        \textbf{IE.Romance  } & it & 1, 1,-1  & 80.9 & 82.0 & 80.3 & -0.6  & 81.8 & 81.4 & +0.5  \\
        \textbf{IE.Romance  } & pt & 1, 1,-1  & 76.8 & 77.5 & 76.0 & -0.8  & 77.4 & 77.6 & +0.8  \\
        \textbf{IE.Romance  } & ro & 1, 1,-1  & 65.8 & 67.7 & 66.3 & +0.5  & 67.7 & 66.9 & +1.1  \\
        \textbf{IE.Romance  } & es & 1, 1,-1  & 74.6 & 75.8 & 74.2 & -0.4  & 75.6 & 74.2 & -0.4 \\
        \textbf{IE.Baltic   } & lv & 1, 1, 1  & 70.3 & 70.7 & 69.5 & -0.8  & 70.5 & 69.9 & -0.4 \\
        \textbf{IE.Latin    } & la & ?, ?, ?  & 47.4 & 48.0 & 45.6 & -1.8  & 48.1 & 47.3 & -0.1 \\
        \textbf{Uralic      } & et & 1,-1, 1  & 65.3 & 65.5 & 65.8 & +0.5  & 65.7 & 66.0 & +0.7  \\
        \textbf{Uralic      } & fi & 1,-1, 1  & 66.7 & 67.1 & 67.0 & +0.3  & 66.9 & 67.1 & +0.4  \\
        \textbf{IE.Germanic } & da & 1, 1, 1  & 76.6 & 76.6 & 76.5 & -0.1  & 76.6 & 76.6 & +0.0  \\
        \textbf{IE.Germanic } & nl & 0, 1, 1  & 67.5 & 67.6 & 67.5 & +0.0  & 67.9 & 67.9 & +0.4  \\
        \textbf{IE.Germanic } & de & 0, 1, 1  & 70.6 & 70.8 & 70.6 & +0.0  & 70.6 & 70.6 & +0.0  \\
        \textbf{IE.Germanic } & no & 1, 1, 1  & 80.5 & 80.4 & 80.5 & +0.0  & 80.5 & 80.5 & +0.0  \\
        \textbf{IE.Germanic } & sv & 1, 1, 1  & 80.3 & 80.5 & 80.5 & +0.2  & 80.5 & 80.5 & +0.2 \\
        \hline\hline
        \multicolumn{2}{l|}{\textbf{Average Performance}} & & 61.1 & 63.8 & 63.2 & +2.2 & 63.6 & 63.1 & +2.0   \\
        \hline
     \end{tabular}
     \caption{Proposed constraints performance compared with baseline, corpus-statistics constraints compiled from WALS features and model utilize the same annotation efforts data.}
     \label{app-table1}
 \end{table*}

\section{Efficientiveness of constraints figure}
\label{app-sec:eff}
\begin{figure}[t!]
            \centering
            \includegraphics[width=\linewidth]{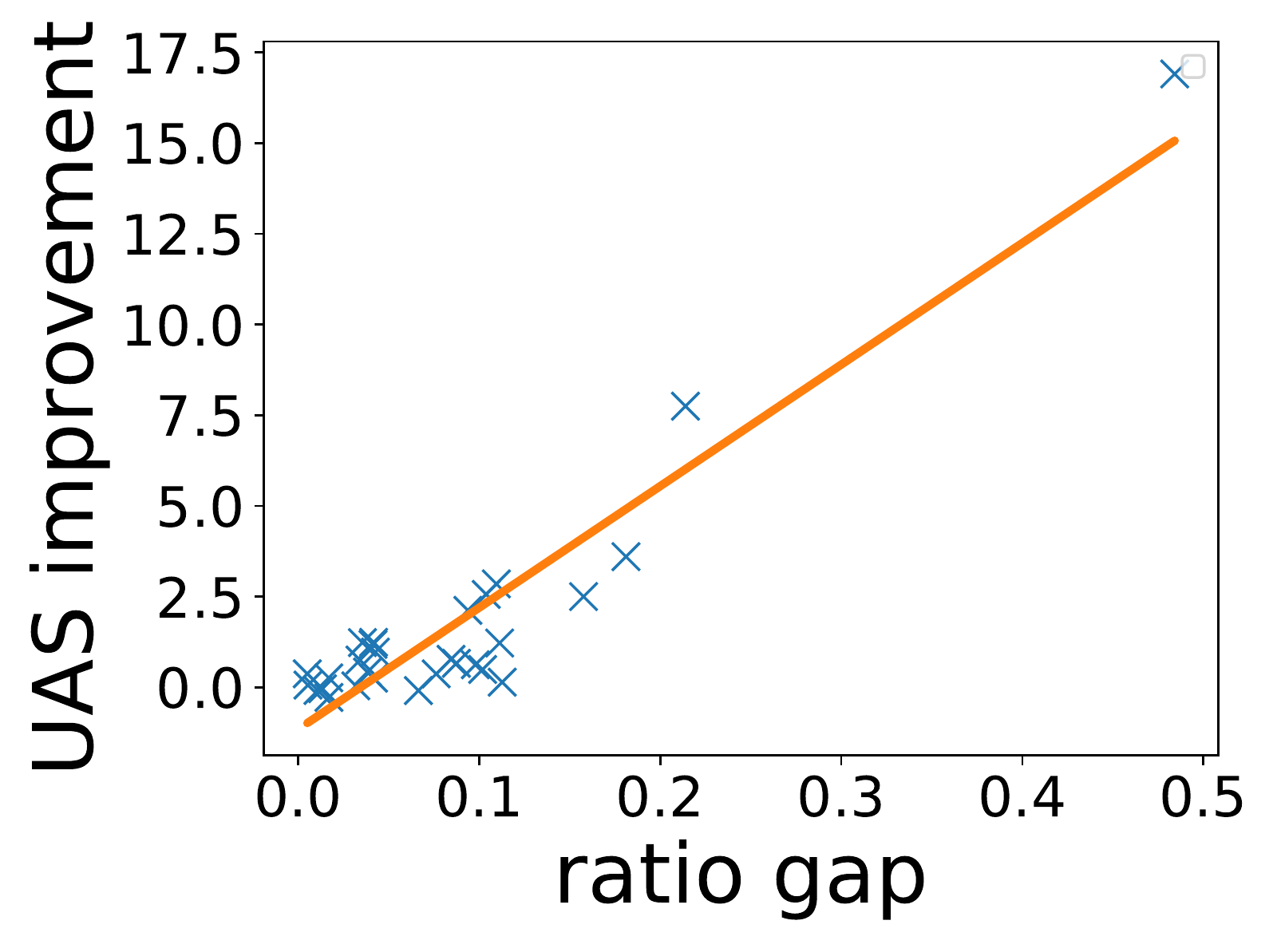}  %
            \caption{Ratio gap v.s. $\Delta$ perf.}
            \label{app-fig:correlation}
            \caption{The performance improvement is highly correlated to the difference in corpus linguistic statistics (estimated by weighted average ratio gaps in constraints) between target and source languages. (The Pearson Correlation Coefficient is $0.938.$)} 
    	\end{figure}

        The figure is shown in Figure \ref{app-fig:correlation}





\end{document}